\documentclass[a4paper, 10pt, conference]{ieeeconf}      % Use this line for a4 paper

\IEEEoverridecommandlockouts                              % This command is only needed if 
\usepackage{graphicx} % for pdf, bitmapped graphics files
\usepackage{subfigure}%for 2 figures in 1 line
\usepackage{verbatim}%for comment
\usepackage{slashbox} %
\usepackage{amsmath}
\usepackage{epstopdf} % for postscript graphics files

\title{\LARGE \bf
Energy Optimization of Automatic Hybrid Sailboat*
}

\author{Ziran Zhang$^{1}$, Zixuan Yao$^{1}$, Qinbo Sun$^{1}$, Huihuan Qian$^{1,2}$% <-this % stops a space
\thanks{*This paper is supported by Project U1613226 from NSFC, Robotic Discipline Development Fund (2016-1418) from Shenzhen Gov, and PF.01.000143 from the Chinese University of Hong Kong, Shenzhen, China.}
\thanks{$^{1}$Ziran Zhang, Zixuan Yao, Qinbo Sun and Huihuan Qian are with The
Chinese University of Hong Kong, Shenzhen.}
\thanks{$^{2}$Corresponding author is
Huihuan Qian, hhqian@cuhk.edu.cn}
}

\begin{document}

\maketitle
\thispagestyle{empty}
\pagestyle{empty}

%%%%%%%%%%%%%%%%%%%%%%%%%%%%%%%%%%%%%%%%%%%%%%%%%%%%%%%%%%%%%%%%%%%%%%%%%%%%%%%%
\begin{abstract}

Autonomous Surface Vehicles (ASVs) provide an effective way to actualize applications such as environment monitoring, search and rescue, and scientific researches. However, the conventional ASVs depends overly on the stored energy. Hybrid Sailboat, mainly powered by the wind, can solve this problem by using an auxiliary propulsion system. The electric energy cost of Hybrid Sailboat needs to be optimized to achieve the ocean automatic cruise mission. Based on adjusted setting on sails and rudders, this paper seeks the optimal trajectory for autonomic cruising to reduce the energy cost by changing the heading angle of sailing upwind. The experiment results validate the heading angle accounts for energy cost and the trajectory with the best heading angle saves up to 23.7\% than other conditions. Furthermore, the energy-time line can be used to predict the energy cost for long-time sailing.

\end{abstract}

%%%%%%%%%%%%%%%%%%%%%%%%%%%%%%%%%%%%%%%%%%%%%%%%%%%%%%%%%%%%%%%%%%%%%%%%%%%%%%%%
\section{Introduction}
The ocean is one of the most precious and valuable resources for the human beings. As the demand increases for marine operations such as environment monitoring [1]-[5], search and rescue [6]-[9], coast preservation [10] and scientific researches [11], Autonomous Surface Vehicles (ASVs) have become a popular research area [1]-[11]. Massive research efforts have been put in this field and most ASVs are catamarans because of stability. Often, they are equipped with only propulsion system such as Wave Adaptive Modular-Vehicle (WAM-V) USV16 from Florida Atlantic University (FAU) [12], Charlie USV from Institute of Intelligent Systems for Automation (ISSIA) [13] and Swordfish ASV from H. Ferreira and his co-workers [14]. They have conducted intensive studies in basic navigation, control and concepts of ASV. Pandey and his co-workers from Osaka University was influenced by WAM-V, and they studied the thrust measurement of the propellers and determined the relationship between the outside force and control force [15].\par
However, considering that wind power is ubiquitous and accessible in the ocean, our laboratory proposed that sailboat, using wind as the main power, is more suitable in marine conditions. Carl Strombeck introduced a modeling and control method to the conventional catamaran sailboat [16].\par 
Nevertheless, the maneuverability of conventional catamarans sailboats is quite low, while ASVs with only propulsion system cannot navigate for a long distance. Inspired by  Cruz's design of catamaran [17], Zhang further designed a Hybrid Sailboat, which combined the advantage of conventional sailboat and propulsion system. By adding the propellers systems to the conventional sailboat, the hybrid system enhances the tacking maneuver and take advantage of the wind power [18]. However, the trade-off is that the electrical energy cost increases. In order to achieve the ocean cruise, the electrical energy cost on the hybrid system needs to be optimized. To address this problem, we first redesign the hybrid power system, which is lighter than the original one, and it is named as Hybrid Sailboat-II.\par
This paper is organized as follows: Section II introduces the mechatronic design of Hybrid Sailboat-II and analyzes the setting of sails and rudders. Section III gives the management of the trajectory. In section IV, the experiment platform is introduced and experiment results are elaborated. The last section concludes the paper.

%---------------------------------------------------------------------------------------------------------
\section{Setting}
\begin{figure}[thpb]
\centering
\subfigure[Hybrid Sailboat-I]
{
\centering
\includegraphics[scale=0.658]{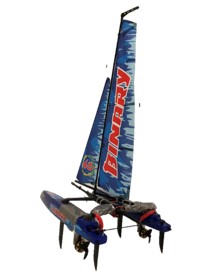}
}
\subfigure[Hybrid Sailboat-II]
{
\centering
\includegraphics[scale=0.33]{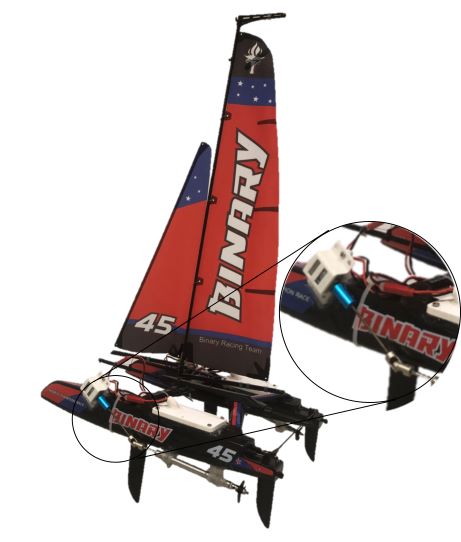}
}
\caption{Assembly layout of Hybrid Sailboat}
\label{sb}
\end{figure}
\subsection{Layout of the Hybrid Sailboat-II}
Hybrid Sailboat-II is a sailboat combined with the propulsion system. Compared to Hybrid Sailboat-I (Fig. \ref{sb}(a)), Hybrid Sailboat-II (Fig. \ref{sb}(b)) adapts lighter Electronic Speed Controllers (ESC) and motors. The space inside the hulls of the sailboat are too narrow to hold two motors inside. Therefore the motors have to be out of the hulls. Two lighter motors are placed on the prow. Cardan joint is introduced to the propulsion system to transmit power from the front motors to propellers instead of using heavy underwater rear motors in the first version. Hybrid Sailboat-II weights 601g, which is 15\% lighter than Hybrid Sailboat-I (691g). Not only is weight reduced, but unbalance is ameliorated. Due to the underwater rear motor, the whole propulsion system of Hybrid Sailboat-I had to be assembled at the back of the hull. Thus, it draws water much deeper at the back than at the front. For Hybrid Sailboat-II, the motor is equipped in front of the hull thus the waterline becomes more balanced. As a result, the sailing performance of Hybrid Sailboat-II is greatly improved. Furthermore, a current-voltage module is equipped to measure and record power consumption during the looping for further analysis of energy optimization.\par
\subsection{PID for Rudders}
Control laws are designed to lead the marine vehicle to reach and follow the desired reference. There are two main components for sailboat: rudders and sails. First, rudders are considered, and a PID regulator is proposed. The chosen control law consists of an algorithm that calculates the necessary rudder angle to reach the desired path in a feasible way.\par
The control law u(t) obtained by a PID controller is given by:
\begin{equation}
u(t)=K_{p}e(t) + K_{i}\int e(t) d\tau +K_{d}\frac{d}{dt} e(t)
\label{p1}
\end{equation}
\begin{equation}
rudder angle = rudder base - pid propotion * u(t)
\label{p2}
\end{equation}
where $e(t)$ is the error signal, which refers to the difference between estimated heading angle and setting angle in this application. Output u(t) is linearly transferred to the necessary rudder angle to reach the desired path, which is illustarted in eq. (\ref{p2}). There are three rudder base angles in our experiment: left base, middle base and right base, corresponding to the the left most rudder angle, middle rudder angle and right most rudder angle. Rudder angle is tuned based on left base during right tacking, based on right base during left tacking, based on middle base in other situations. Additionally, pid propotion can be adjusted in order to achieve a radical or conservative rudder policy. Moreover, when applying the PID rudder control, a clipping idea is adapted. u(t) is amplified linearly so that maximum rudder angle $40^{\circ}$ is maintained when turning a large maneuvering angle.\par
\begin{figure}[thpb]
\centering
\includegraphics[scale=0.045]{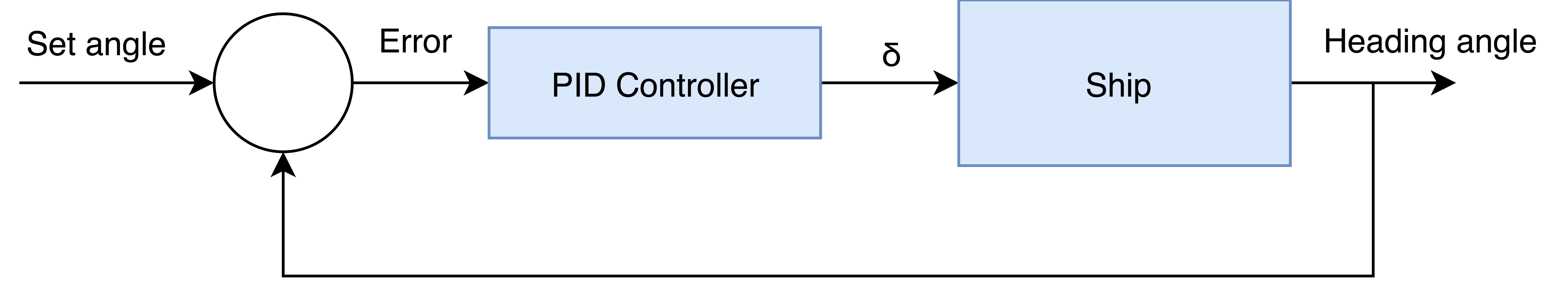}
\caption{The PID conception applied to Hybrid Sailboat-II}
\label{pid}
\end{figure}
When the rudder torque and wind torque achieved equilibrium, the heading angle is not exactly the desired angle. Due to equilibrium state, $K_{p}$, $K_{d}$ are unable to further change the angle. However, this error will be accumulated by integral and change the rudder angle to reach desired heading eventually.\par
\begin{figure}[thpb]
\centering
\subfigure[Design of test]
{
\centering
\includegraphics[scale=0.084]{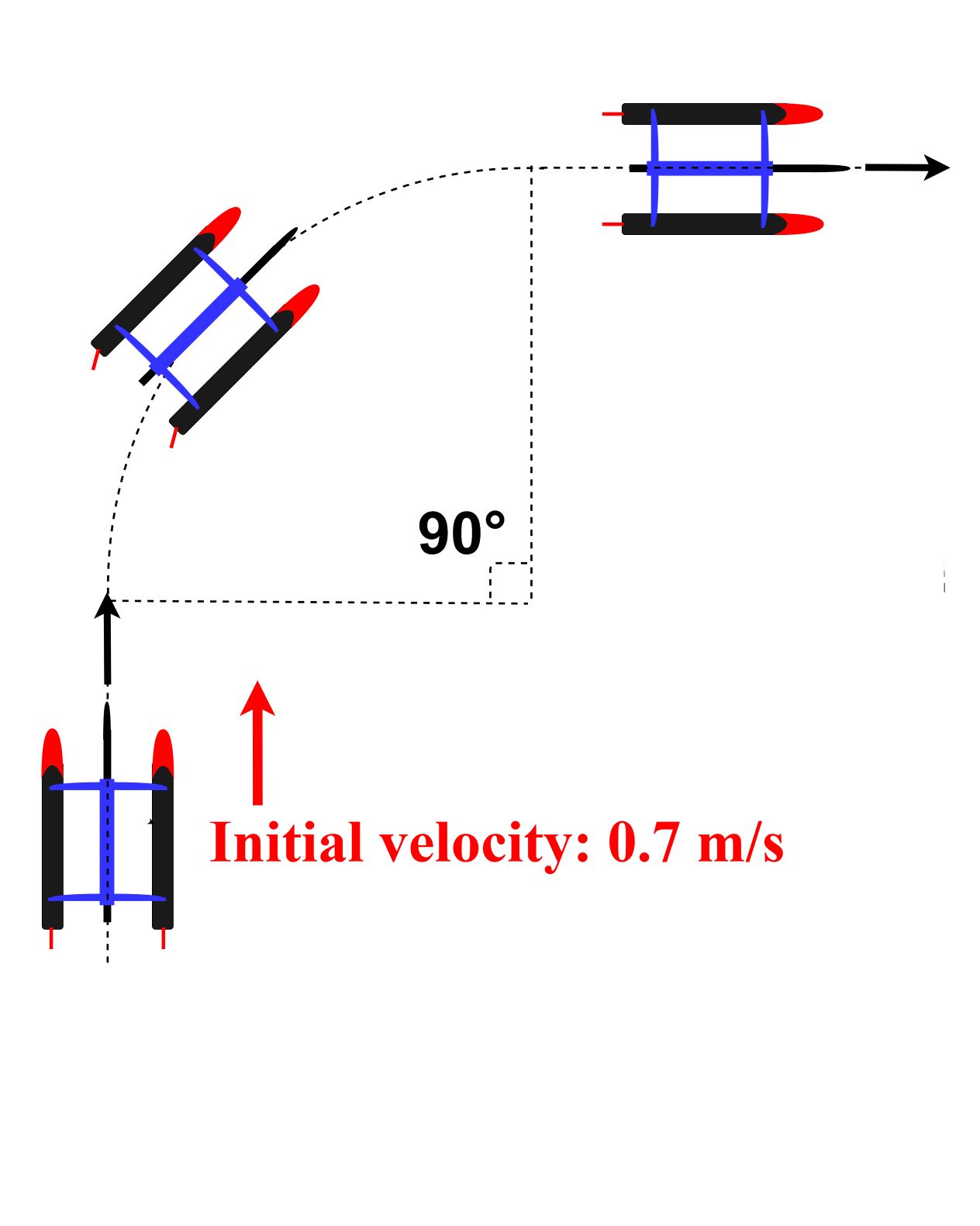}
}
\subfigure[Test result]
{
\centering
\includegraphics[scale=0.35]{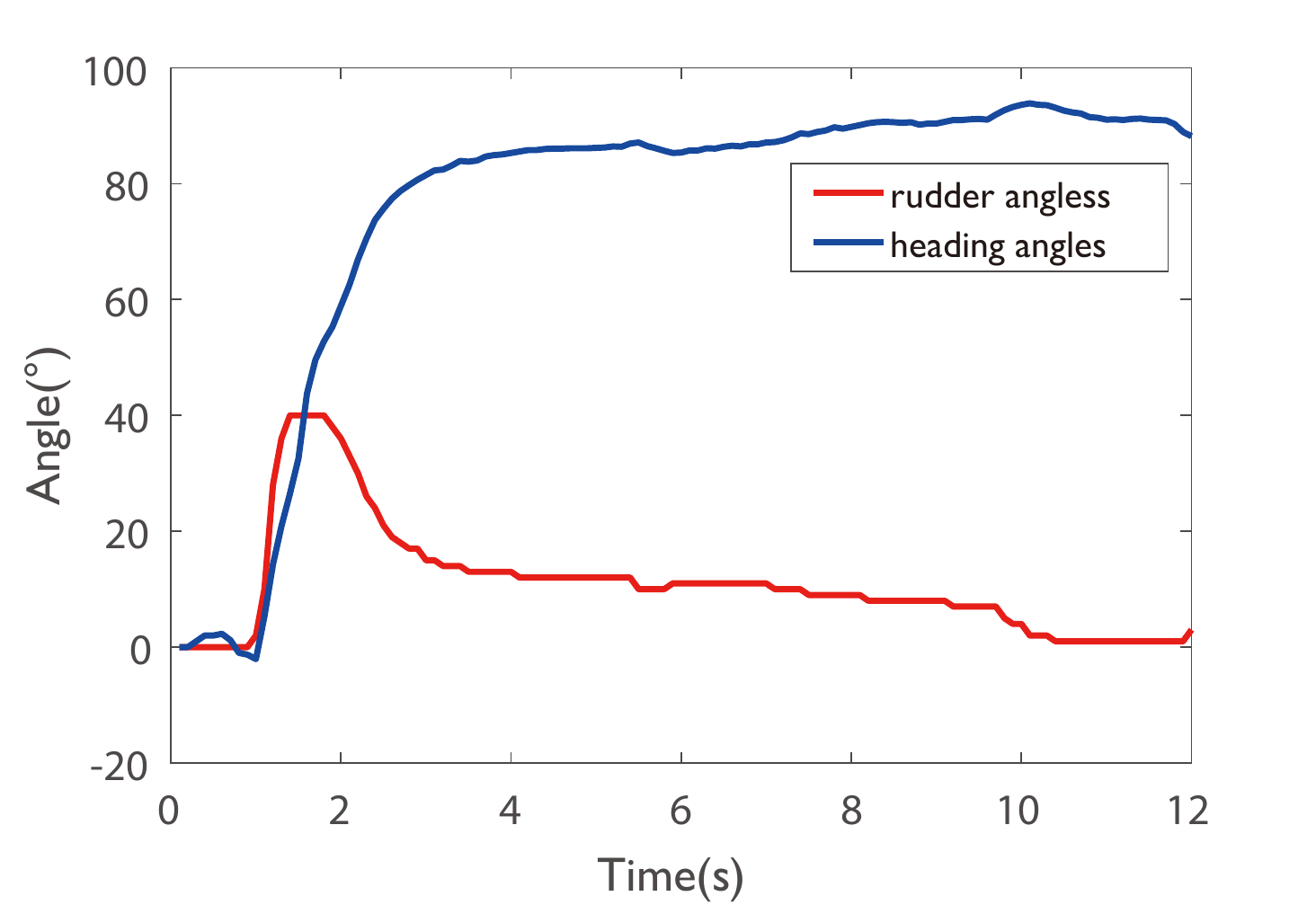}
}
\caption{The PID test on Hybrid Sailboat-II rudders}
\label{pida}
\end{figure}

Generally, a trade-off exists between speed and stability for different sets of PID parameters. PID parameters, which can reach the desired path quickly, will overcorrect and vibrate before the path is finally stable; while smooth and stable path could be achieved in sacrifice of the speed. It is more significant for the sailboat to be stable than to be maneuverable when turning in the wind field because the propellers will further help the sailboat to maneuver. Thus, the PID parameters are tuned, $K_{p}=0.2$, $K_{i}=0.1$ and $K_{d}=0.01$ are tested to be suitable. The sailboat is released manually with a speed approximately to sailing (0.7m/s) and the difference between the released heading angle and the desired heading angle is  $90^{\circ}$ (Fig. \ref{pida}(a)). In the Fig. \ref{pida}(b), the heading angle will reach $80^{\circ}$ in a short time and then be adjusted slowly and smoothly to $90^{\circ}$.\par

Wind is the only source for propulsion for a sailboat in the ocean. Therefore, the sail should be modeled to provide optimal force for better motion control. Based on Sun's previous work [19], the optimal angle of the sail can be determined.\par
\begin{figure}[thpb]
\centering
\includegraphics[scale=0.13]{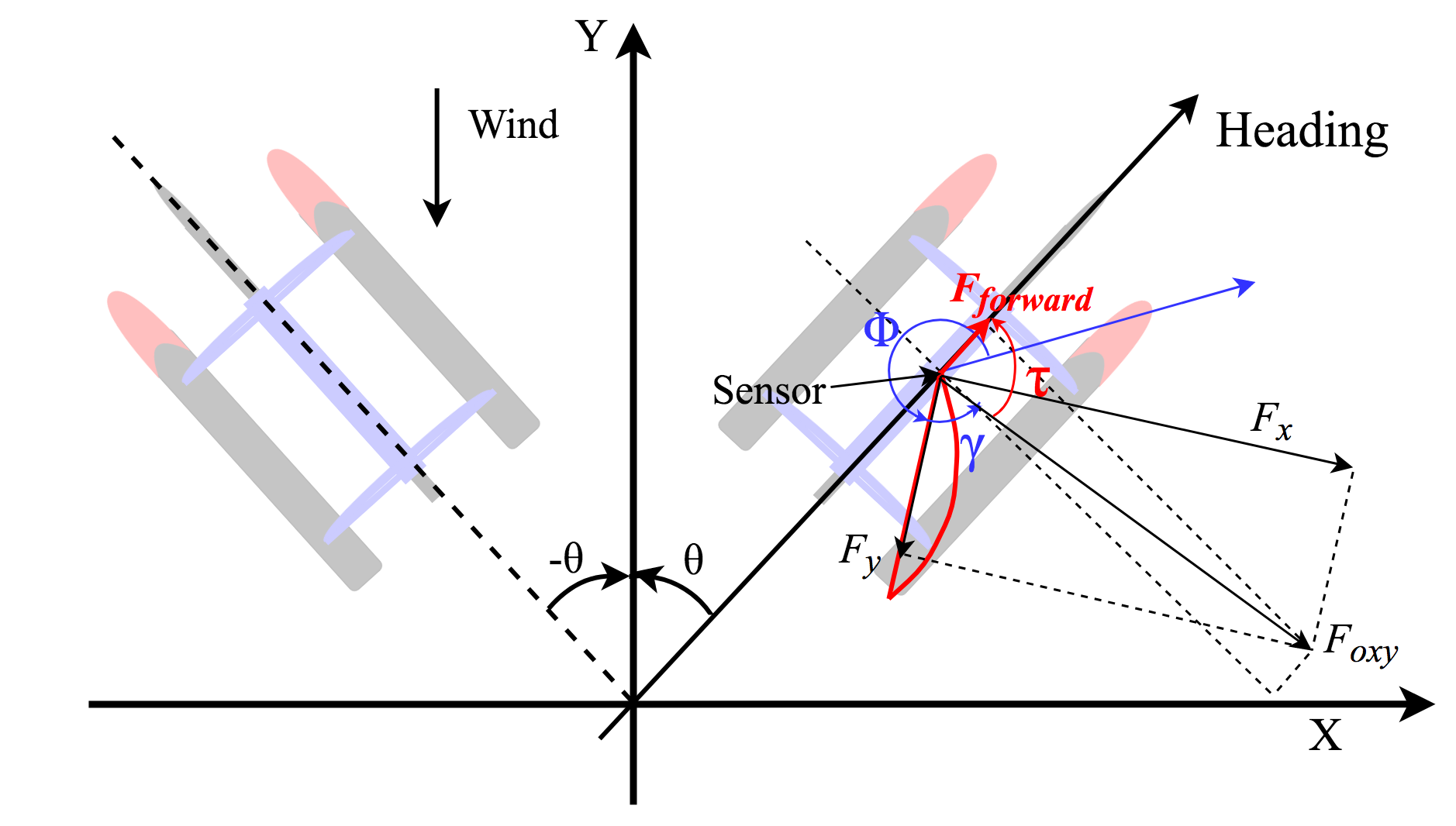}
\caption{Analysis of forces and angles}
\label{wforce}
\end{figure}

As Fig. \ref{wforce} shows, a sensor is fixed on the sail, which can measure the force that the wind exerts on the sailboat. The sensor can measure the force in two orthogonal directions, which are exhibited as $F_{x}$ and $F_{y}$ and these two directions will rotate with the sail. The sail angle is defined as $\Phi \in [0,360]$ in the angular coordinate. The combined forces of $F_{x}$ and $F_{y}$ is denoted as $F_{oxy}$.\par
\begin{figure}[thpb]
\centering
\subfigure[$F_{right-tacking}$]
{
\centering
\includegraphics[scale=0.255]{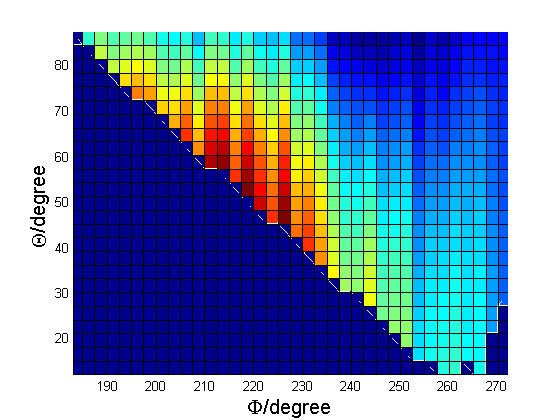}
}
\subfigure[$F_{left-tacking}$]
{
\centering
\includegraphics[scale=0.255]{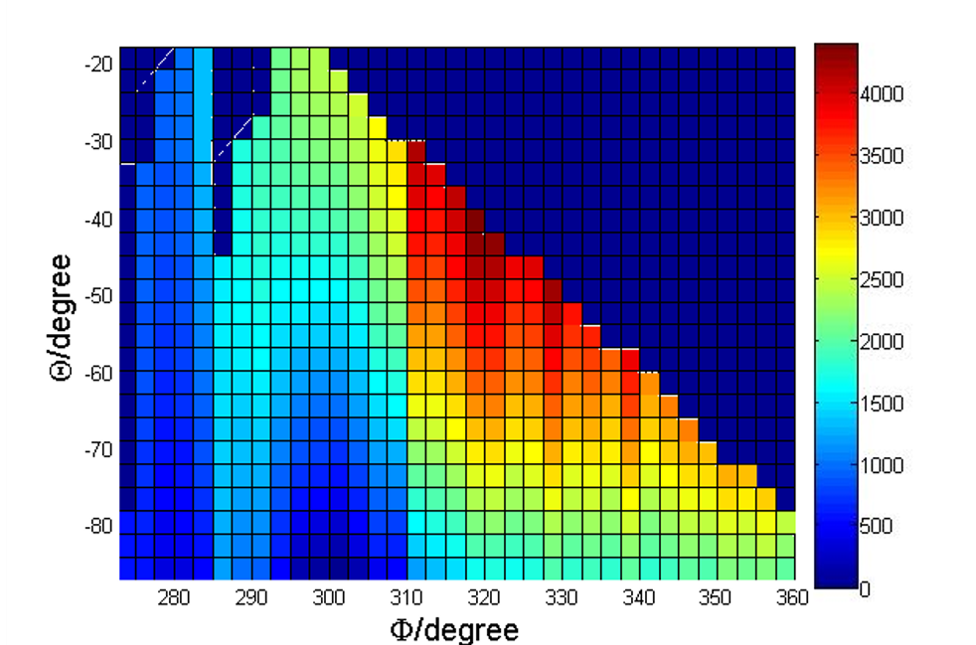}
}
\caption{Forward force distribution map}
\label{wa}
\end{figure}

To decompose the $F_{oxy}$ in the heading direction, we define:
\begin{equation}
F_{forward}=| \mathbf{\bar{F}_{oxy}} |\times \cos \tau 
\label{s1}
\end{equation}
\begin{equation}
\begin{aligned}
\tau=\frac{5\pi}{2}-\theta-\Phi-a\tan2(F_{x},F_{y})
\end{aligned}
\label{s2}
\end{equation}
After various experiments, a force distribution map is plotted to show the magnitude of $F_{forward}$ for each pair of $(\theta,\Phi)$ as shown in Fig.\ref{wa}. Magnitude of $F_{forward}$ is represented by a color bar where stronger force is in red and weaker in blue. Fig.\ref{wa}(a) is for tacking towards right and Fig.\ref{wa}(b) is for tacking towards left. Referring to this map, we can locate the optimal angular coordinate $\Phi$ for a given heading angle $\theta$, which is then converted into a desired sail angles with respect to the hull of the sailboat.\par

%---------------------------------------------------------------------------------------------------------
\section{Trajectory Planning}
\begin{figure}[thpb]
\centering
\subfigure[WRSC sailing trajectory]
{
\centering
\includegraphics[scale=0.46]{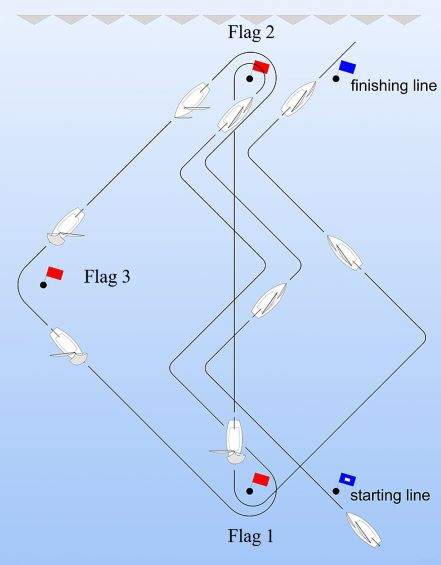}
}
\subfigure[Cruise trajectory design for Hybrid Sailboat-II]
{
\centering
\includegraphics[scale=0.06]{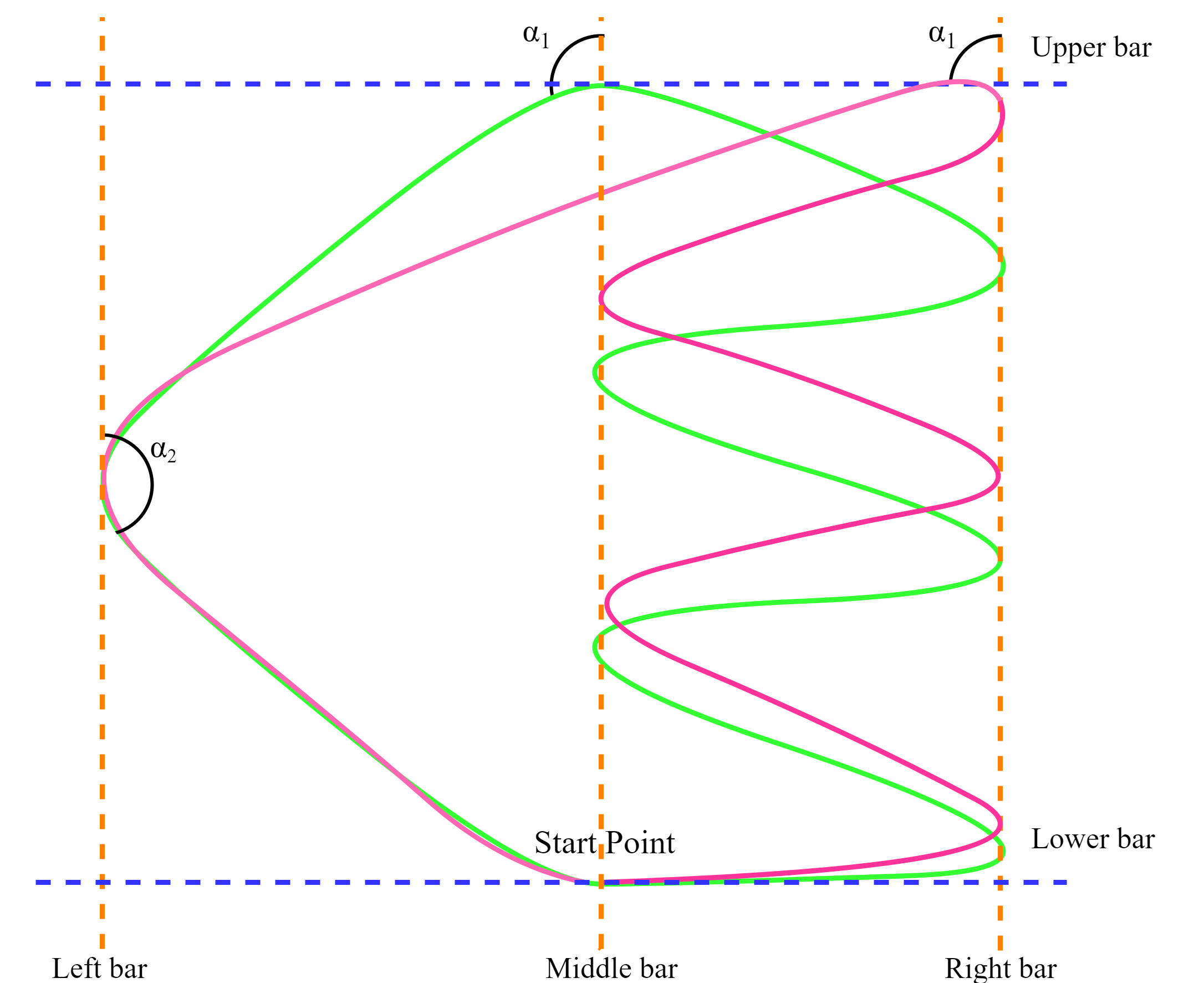}
}
\caption{Sailing trajectory}
\label{wrsc}
\end{figure}

Our experimental track refers to World Robotic Sailing Championship (WRSC), as Fig.\ref{wrsc}(a) shows. Due to the limitation of our experiment platform size and the requirement of automatic cruising (looping), we only select the left half of the WRSC track, which is confined by the three red flags. In our trajectory design, we set four bars to limit this reigon shown as Fig.\ref{wrsc}(b), and they are named as left bar, right bar, upper bar,and lower bar.\par

Based on our main idea of energy-saving, some further adjustments of the track are made. At flag 1 of Fig \ref{wrsc} (a), the Hybrid Sailboat-II need to make a turn of $180^{\circ}$ with barely any effective displacement. It will consume a lot of energy to make this turn. Therefore, we decided to let Hybrid Sailboat-II first sails beam reach when it returns to the start point, as shown in Fig \ref{wrsc} (b). During this stage, the boat will consume much less energy for the motors are all off. The Hybrid Sailboat-II will start to sail close-hauled when it first reaches the right bar. Apparently, this behavior takes less energy because the turning angle is less than $180^{\circ}$, which means the propeller will be on for a shorter period. For the Hybrid Sailboat would reach the upper bar during both sailings right close-hauled or left close-hauled due to different heading angle $\theta$, there are two possible tracks under our track design, shown in Fig.\ref{wrsc}(b). The green one shows the Hybrid Sailboat sails back when it reaches the upper bar during left close-hauled while the pink one shows Hybrid Sailboat-II sails back when it reaches the upper bar during right close-hauled. \par

Another bar named as middle bar is set in the middle of the region. The Hybrid Sailboat will do tack in the region between the middle bar and right bar. When the boat is heading right (heading angle = $\theta$) and reaches the right bar, the setting angle will be changed to $-\theta$, as shown in Fig.\ref{wforce}. When the boat is heading left (heading angle = $-\theta$) and reaches the middle bar, the setting angle will be changed to $\theta$ again. During tacking, one side of the motors are turned on to assist tack when the Hybrid Sailboat-II needs to change its heading, and the motor will be turned off when the angle difference is reduced to a boundary angle for saving energy. In other situations, only rudders and sails are effective. The sailboat will reach the target angle by inertia and the bound angle is tested in our previous experiments to be $30^{\circ}$. Therefore, the Hybrid Sailboat-II can do tack in the left region. When the boat reaches the upper bar, the setting angle will be changed to $-\alpha_{1}$ (Fig. \ref{wrsc}(b)), which is calculated by the current position and midpoint of the left bar. Then, when the boat reaches the left bar, the setting angle will be changed to $\alpha_{2}$, which is calculated by the current position and start point. Therefore, Hybrid Sailboat-II will return to the start position and be able to start another loop. \par

In the above design, $\theta$ is changeable, and various $\theta$ will be tried to determine the relationship between $\theta$ and total energy consumption during cruising in the area determined by 4 peripheral bars.

% \begin{figure}[thpb]
% \centering
% \includegraphics[scale=0.08]{tackingc}
% \caption{Cruise trajectory design for Hybrid Sailboat-II}
% \label{tackc}
% \end{figure}

%\begin{figure}[thpb]
%\centering
%% \subfigure[Heading left]
%%{
%% \centering
%\includegraphics[scale=0.07]{boatl}
%%}
%% \subfigure[Heading right]
%%{
%%\centering
%% \includegraphics[scale=0.32]{boatl}
%%}
%\caption{Sailboat heading angle defination}
%\label{bt}
%\end{figure}

%---------------------------------------------------------------------------------------------------------
\section{Experiment}
We assume that there are two main reasons that the heading angle can affect tacking efficiency. The first reason is that the tacking distance varies for different heading angle. The farther one tack goes, the less number of tackings is needed for given distance. The second one is that different heading angles will result in different tacking maneuver angles (equal to $2\theta$). The smaller the tacking maneuver angle is, the less energy will be consumed by motors. Therefore, the optimal heading angle needs to satisfy both long tacking distance and small tacking maneuver angle. The experiment is designed to find this optimal heading angle to minimize the total energy consumption in automatic cruising. Hybrid Sailboat-II will follow the trajectory shown in section III and sail for 5 loops with various heading angles $\theta$. The total energy consumption is compared to seek the heading angle consuming the least energy. \par
\subsection{Experiment Setup}

\begin{figure}[thpb]
\centering
\includegraphics[scale=0.40]{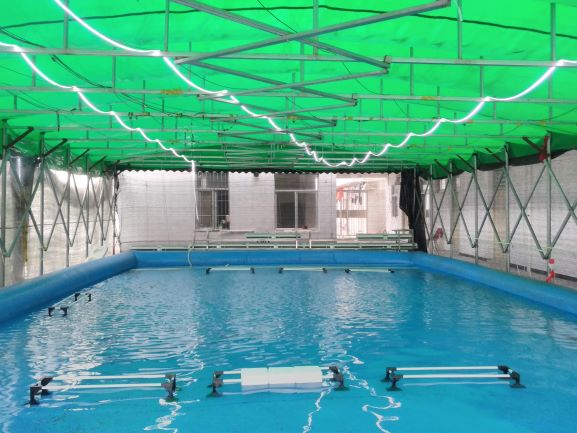}
\caption{The experiment platform (STAr) for ASVs}
\label{star}
\end{figure}

The experiment platform is mainly based on the Sailboat Testing Arena (STAr). Fig. \ref{star} gives the layout of the platform, it consists of an 8m$\times$12m water pool with wind field and a roof with 6 cameras. The velocity of the wind field provided by the fans is 1.2-1.4 m/s. The capture area of the 6 separate cameras are combined together by a CV algorithm, and every point under the camera area is assigned to a specific coordinate (x, y). In addition, STAr can distinguish the color of Hybrid Sailboat-II, and extract its pixel coordinate as long as it is under the camera area.\par

Hybrid Sailboat-II uses a Raspberry Pi to receive commands from PC and send current status of the boat back to PC, including heading angle $\theta$, rudder angles, sail angles and motor information. Combining all these status information with the position from STAr, the control program in PC will determine the next action of the sailboat. Hybrid Sailboat-II will receive the command and cruise in the pool automatically.\par 

The range of heading angle $\theta$ is selected from $35^{\circ}$ to $55^{\circ}$, because that is the range of sailing close-hauled for most sailboats and it will be verified by the experiment that there is no need to explore other angles beyond this bound. $\theta$ is changed by $5^{\circ}$ every time, if the difference is less than $5^{\circ}$, the energy change is trivial. The Hybrid Sailboat-II is released at the same point of the pool and the battery voltage is controlled within the range of 7.8-8.4v and other settings of the experiment platform remained unchanged.\par

\subsection{Experiment Results \& Analysis}
\begin{figure}[thpb]
\centering
\subfigure[Heading angle=55$^{\circ}$]
{
\centering
\includegraphics[scale=0.2]{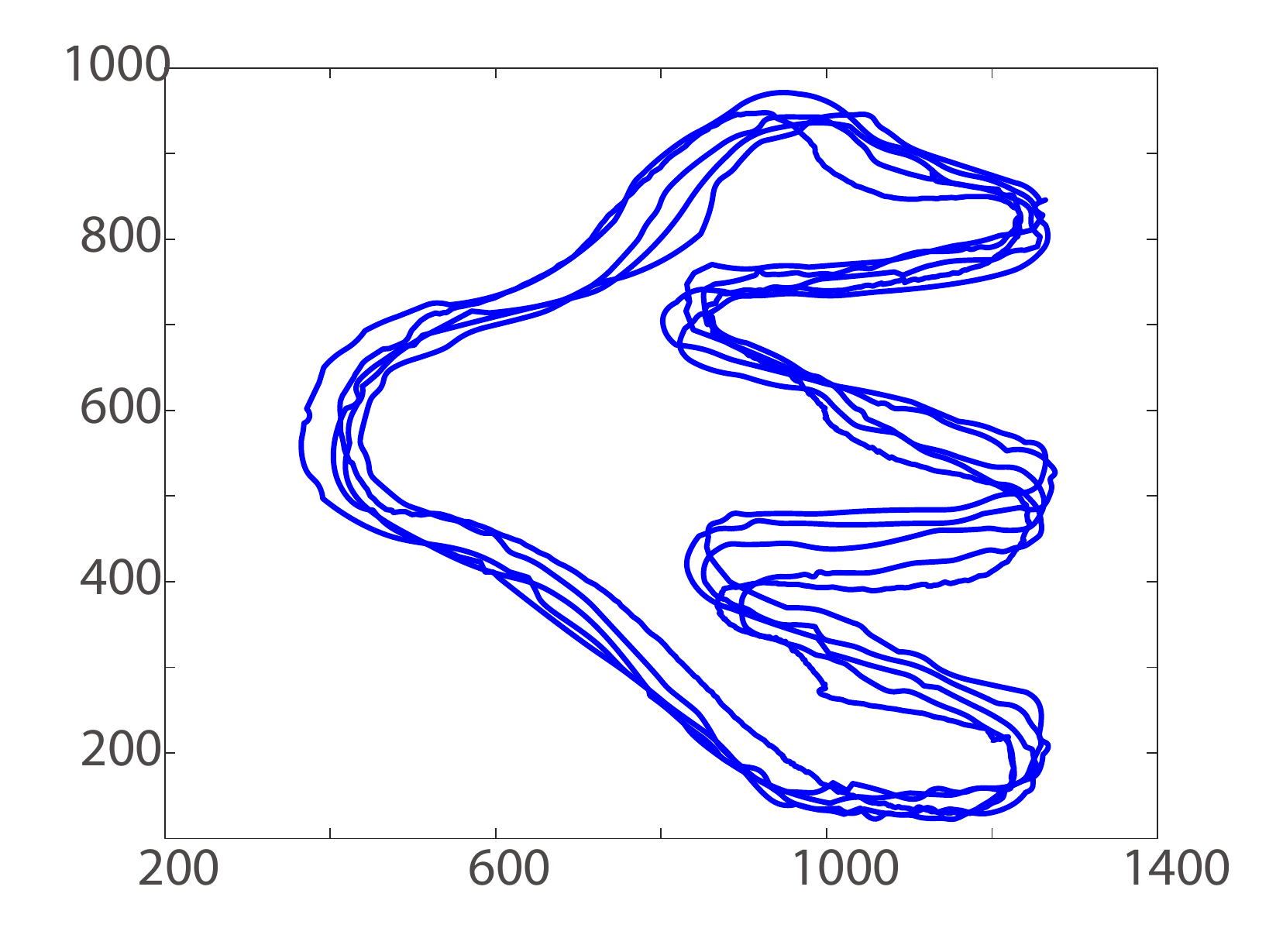}
}
\subfigure[Heading angle=50$^{\circ}$]
{
\centering
\includegraphics[scale=0.2]{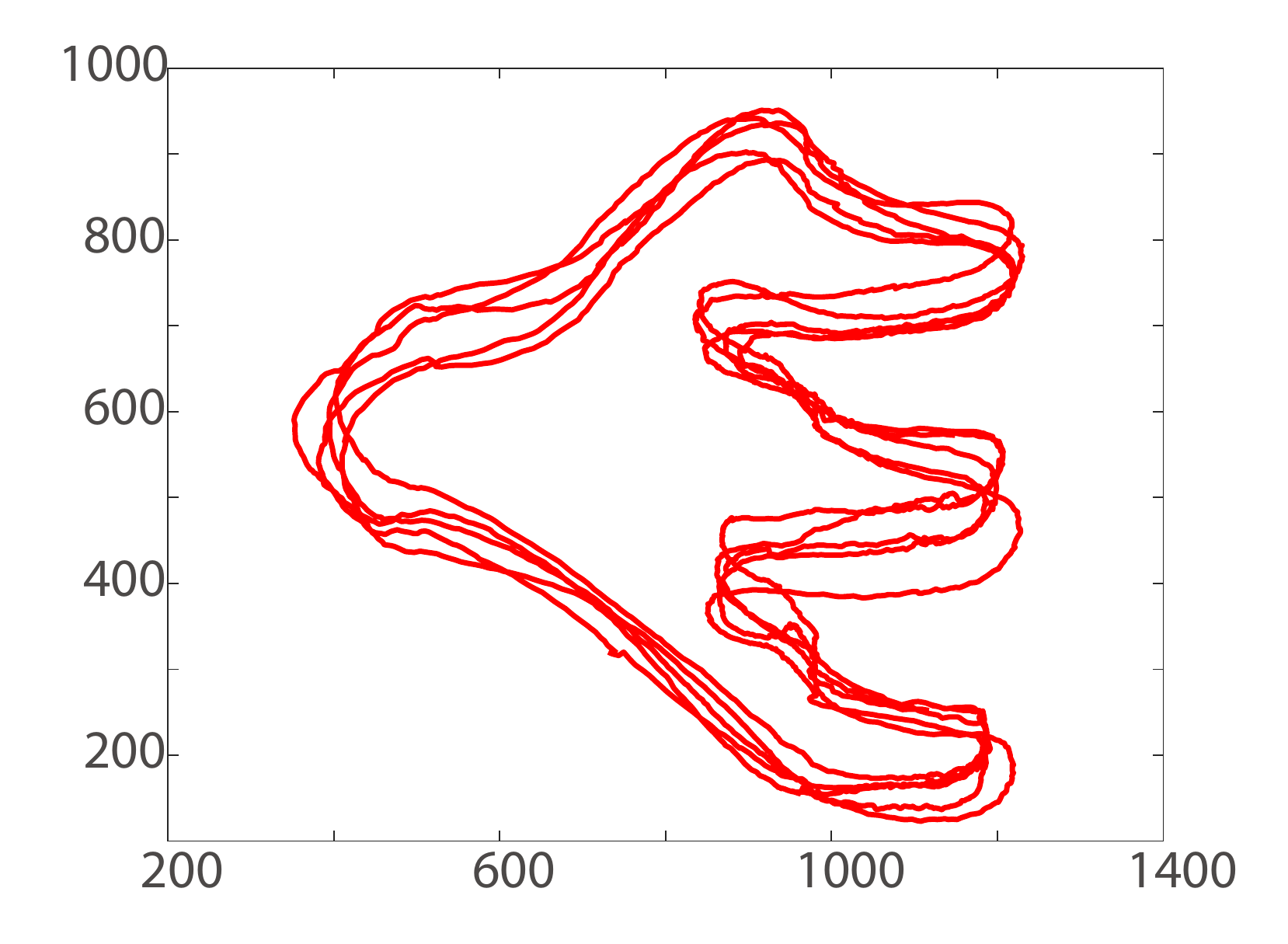}
}
\subfigure[Heading angle=45$^{\circ}$]
{
\centering
\includegraphics[scale=0.2]{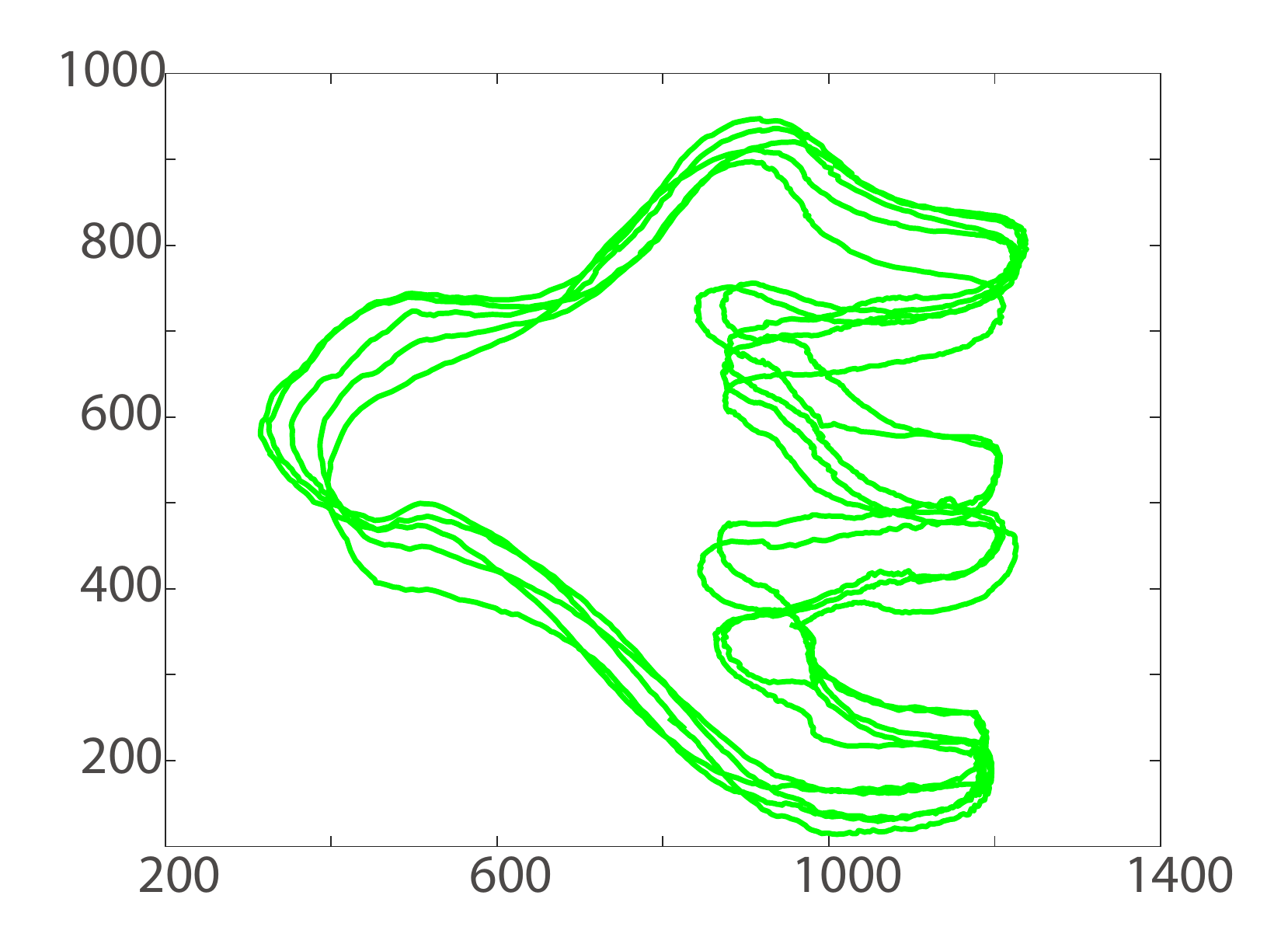}
}
\subfigure[Heading angle=40$^{\circ}$]
{
\centering
\includegraphics[scale=0.2]{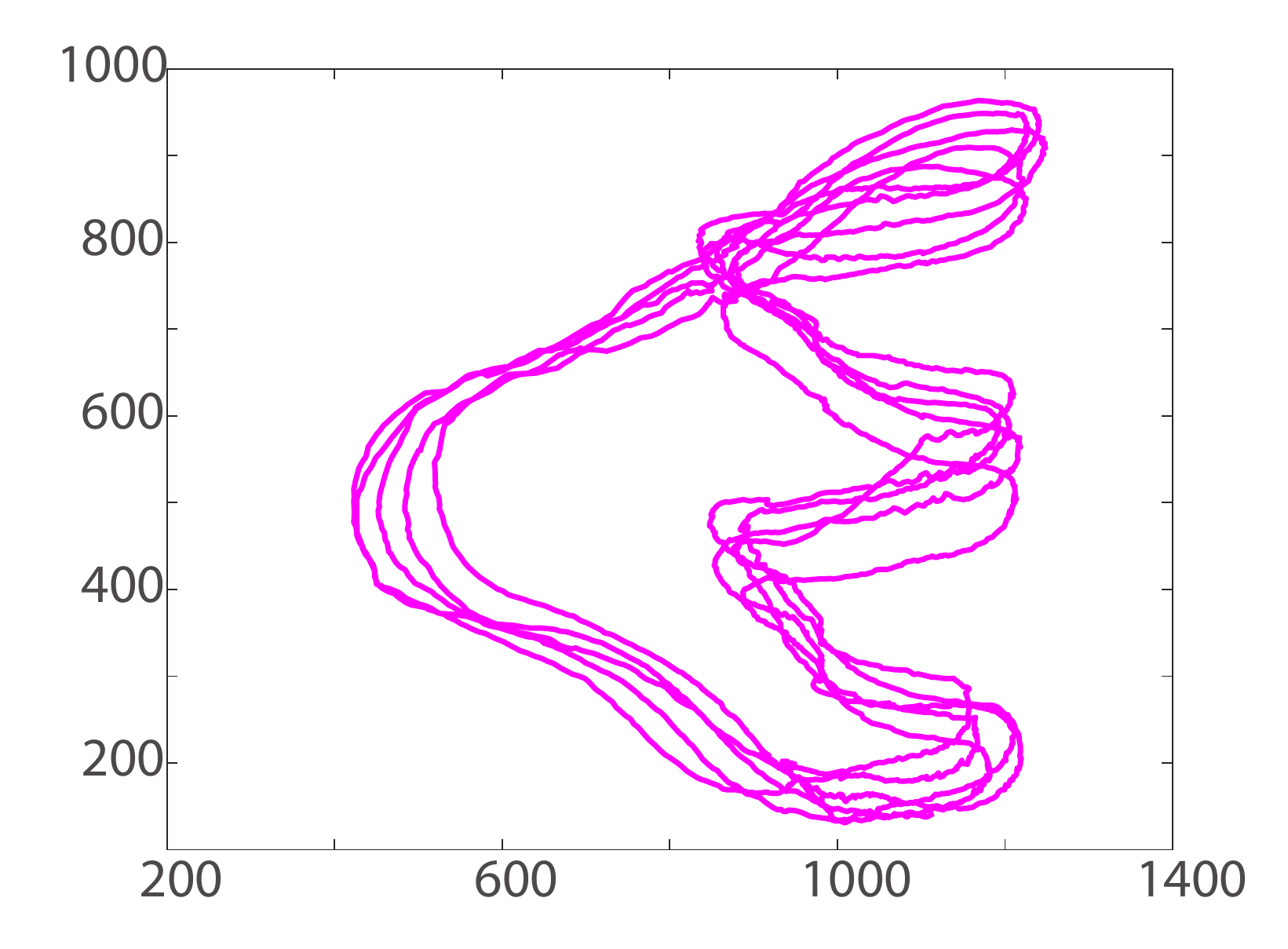}
}
\subfigure[Heading angle=35$^{\circ}$]
{
\centering
\includegraphics[scale=0.2]{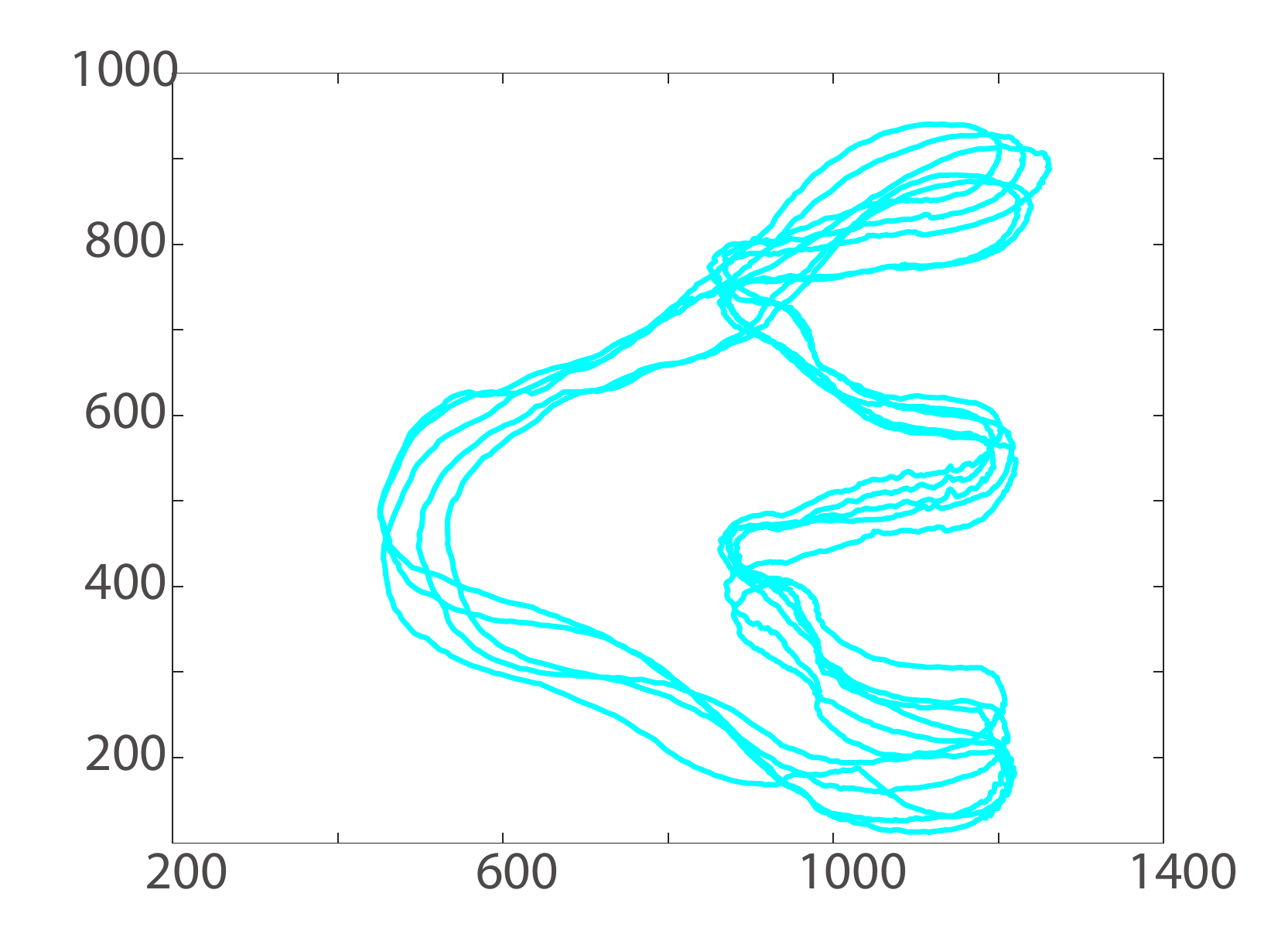}
}
\caption{Trajectories for different heading angles}
\label{five}
\end{figure}

\begin{figure}[thpb]
\centering
\includegraphics[scale=0.5]{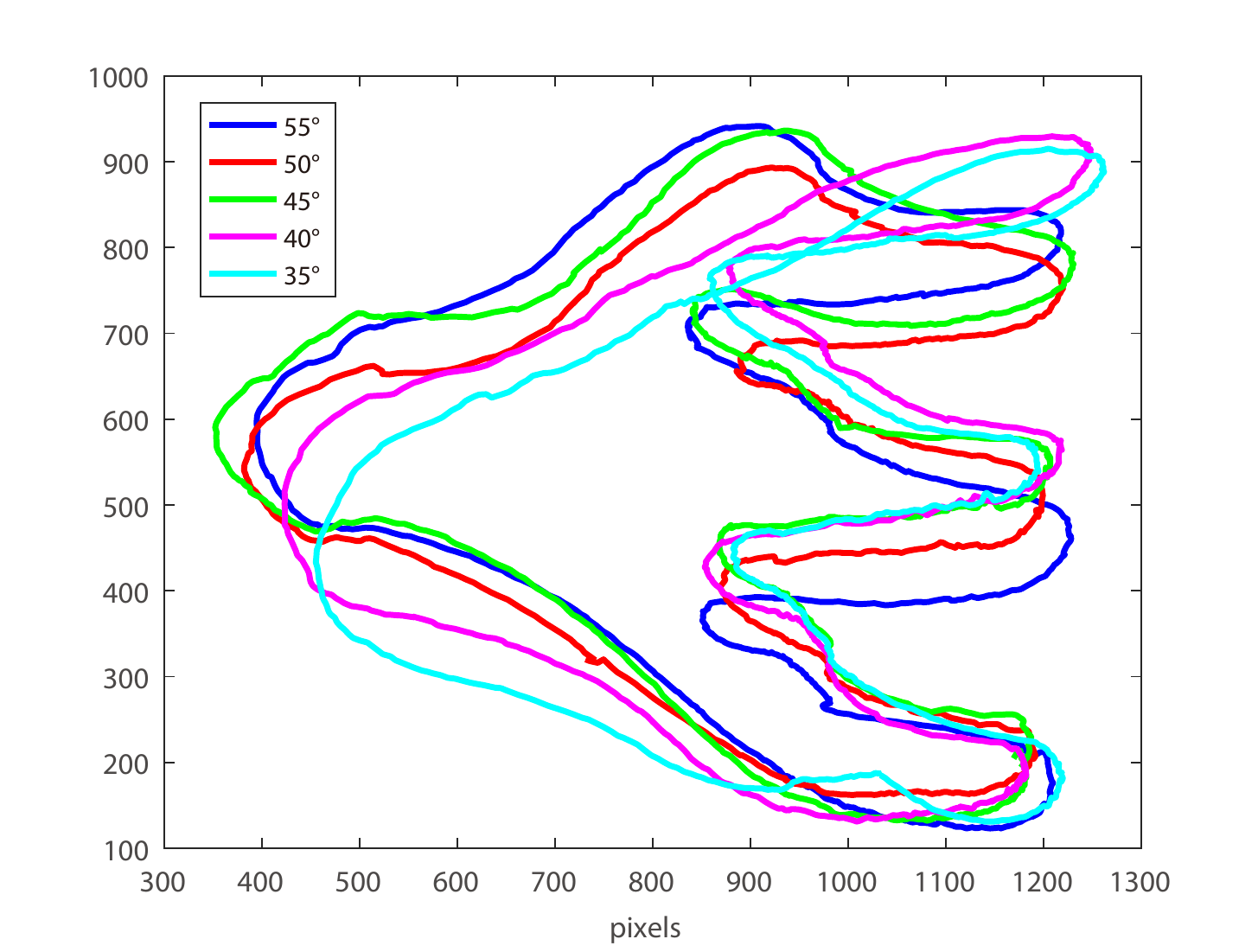}
\caption{The loop trajectories for different heading angles}
\label{traj}
\end{figure}

Fig. \ref{five}(a)-(e) shows the general trajectories with $\theta$ from $35^{\circ}$ to $55^{\circ}$ respectively. In each subfigure of Fig. \ref{five}, every loop is similar to other loops with the same $\theta$, and it can be illustrated that Hybrid Sailboat-II is well-controlled and every loop is representative. One loop is selected from every group of the experiment to study regularity. The trajectories are compared, as shown in Fig. \ref{traj}. For the left part of the figure, the left bounds of these loops are not the same because sometimes the board reach speed is quite fast and the sailboat would dash out the setting bound. For the right part of the figure, the deviation angles of trajectories are not exactly the same as $\theta$ because the sailboat actually drifts when beating up to windward. Concentrating on a single loop, the heading angle of blue loop ($55^{\circ}$) is too large so that it is very hard for Hybrid Sailboat-II to keep its heading direction due to the large wind torque. Thus, the blue trajectory is not smooth. For the cyan trajectory, the sailboat sometimes stops during close-hauled period because its heading angle reaches the no-go zone. It can be concluded that $35^{\circ}$ (cyan) and $55^{\circ}$ (blue) are the limits of $\theta$, and there is no need to explore heading angles smaller than $35^{\circ}$ and larger than $55^{\circ}$. \par

\begin{figure}[thpb]
\centering
\includegraphics[scale=0.5]{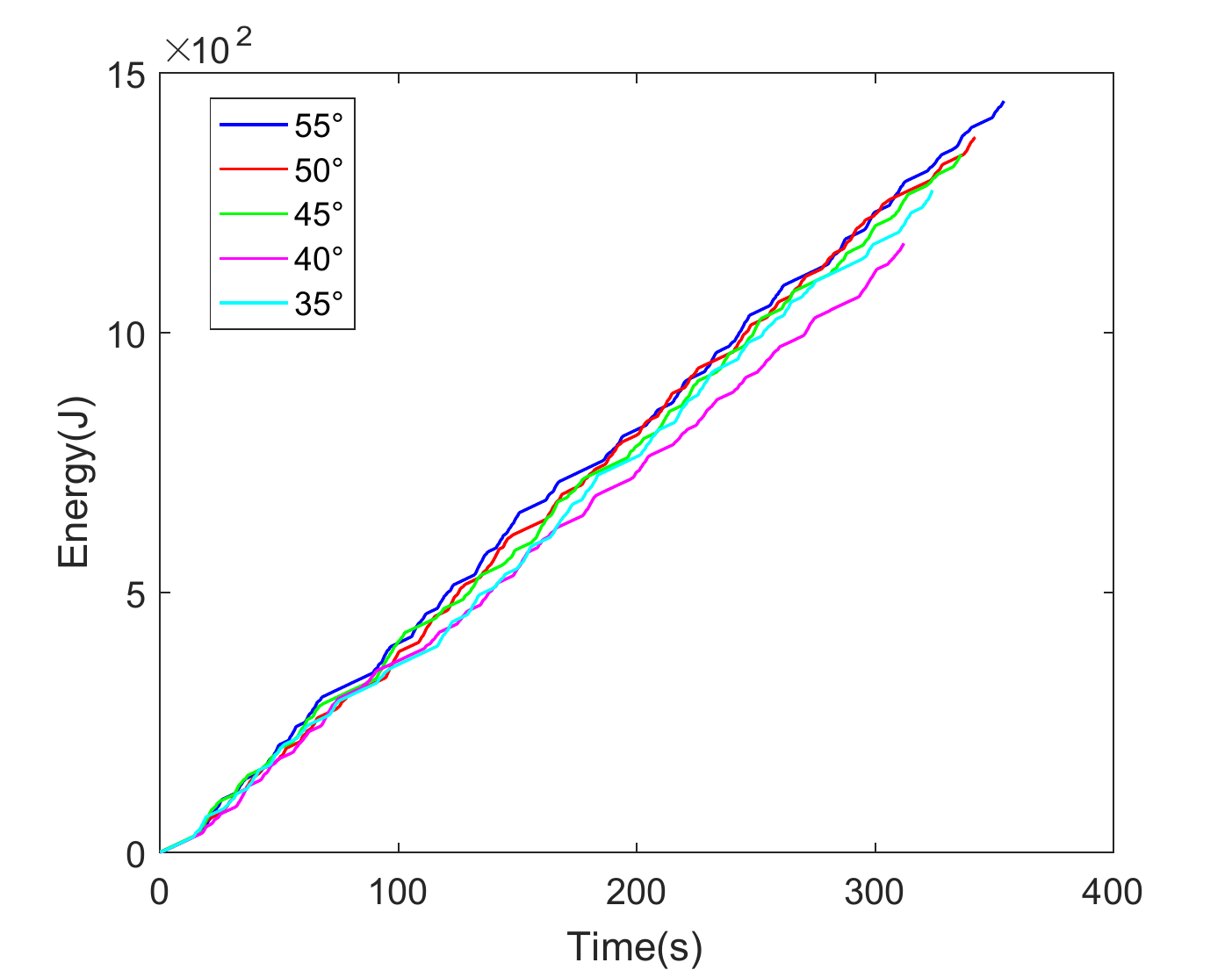}
\caption{Integration for energy cost of different heading angles}
\label{ener}
\end{figure}

\begin{figure}[thpb]
\centering
\includegraphics[scale=0.45]{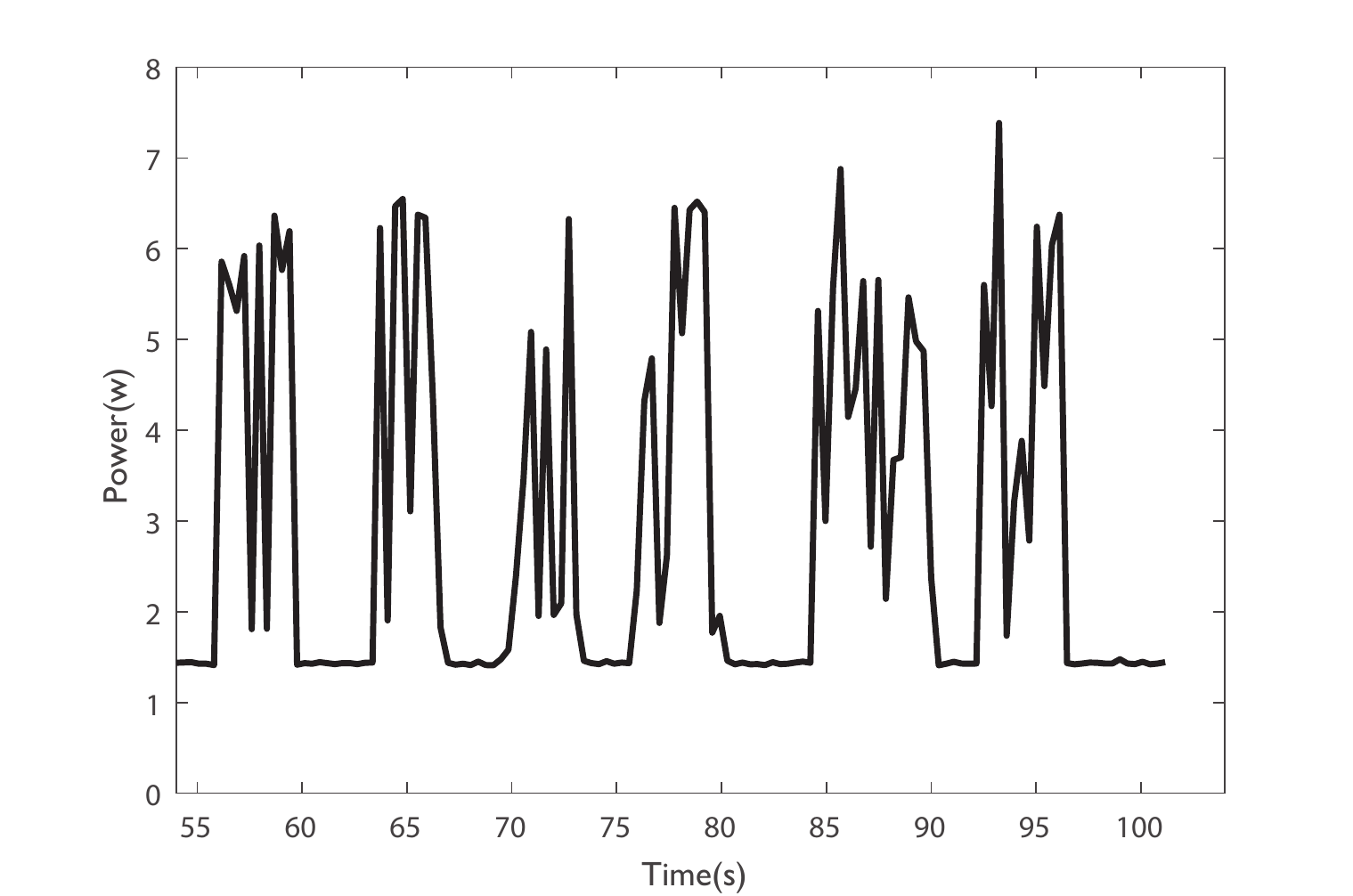}
\caption{The power consumption for one sample loop, captured from the situation heading angle equals $40^{\circ}$}
\label{powerl}
\end{figure}

\begin{figure}[thpb]
\centering
\subfigure[Energy per loop-Heading angle]
{
\centering
\includegraphics[scale=0.235]{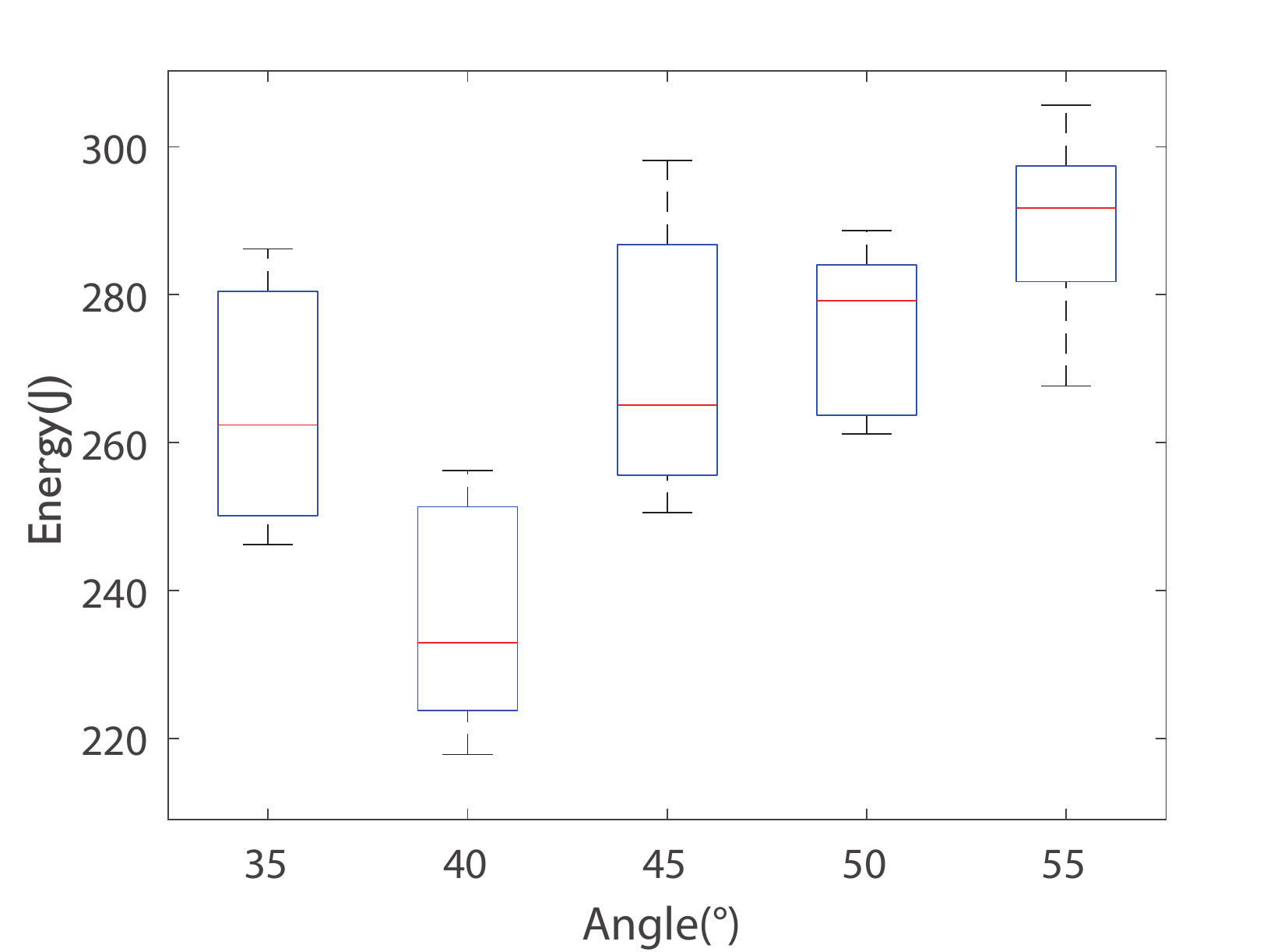}
}
\subfigure[Tacking energy-Heading angle]
{
\centering
\includegraphics[scale=0.235]{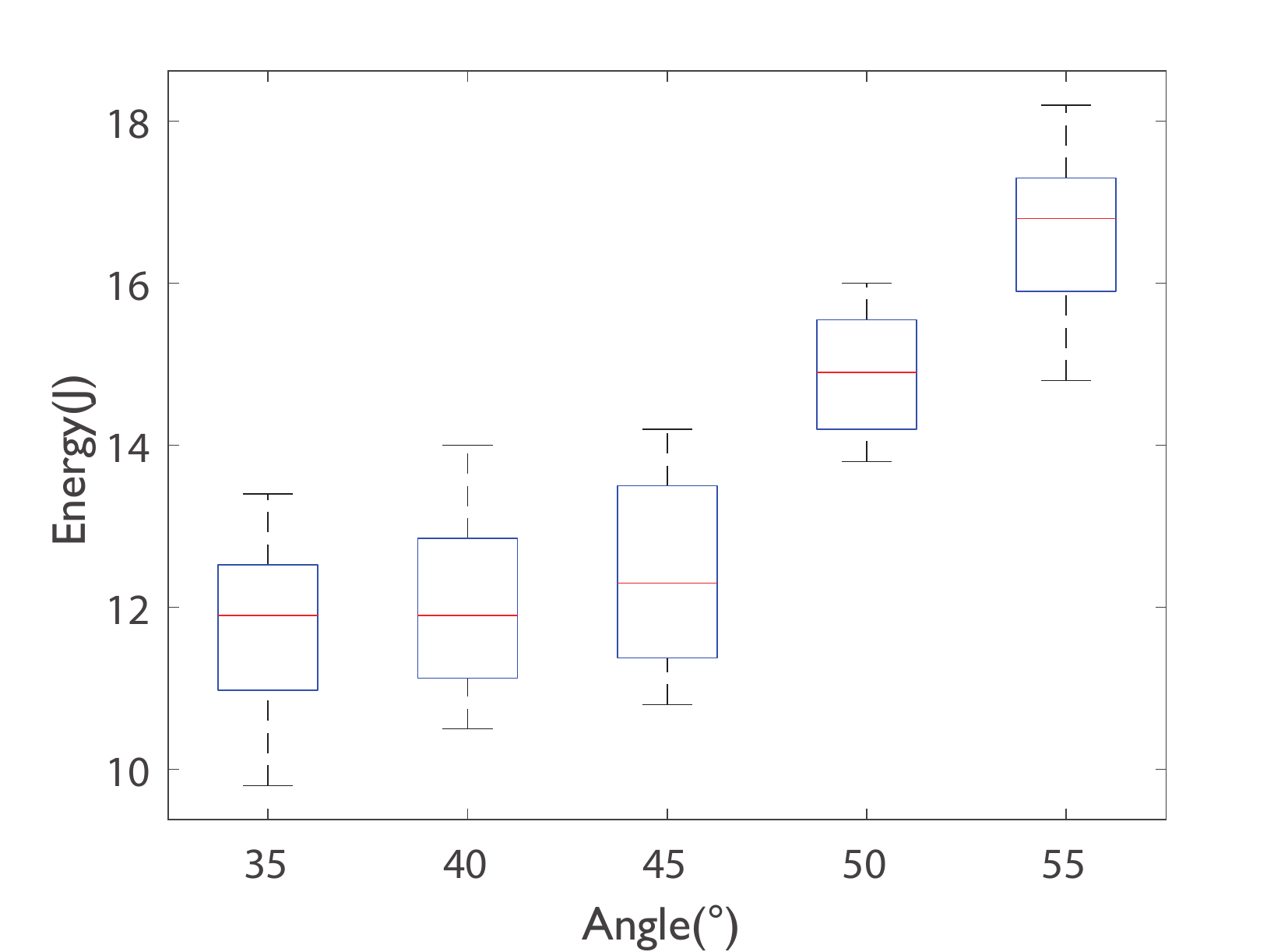}
}
\caption{Energy cost for different heading angles}
\label{enpic}
\end{figure}

Then, the energy consumption of each loop with different $\theta$ is exhibited in Fig. \ref{ener}, which is actually taking the integration to sum up the total energy consumption of 5 loops. This figure shows the relationship between the total energy consumption of Hybrid Sailboat and time, and the color of this figure is consistent with the color of Fig. \ref{traj}. The lengths of these lines are not the same, because the time which Hybrid Sailboat-II taskes to finish 5 loops varies. The slopes of the lines reflect the speed of energy cost. The sudden increase of the slope represents the using of motors. Fig. \ref{powerl} shows the power consumption of one sample loop. In this figure, the valley reflects the power of all devices with motors off and the peak value means the total power including open motors. Some lines cross each other in Fig. \ref{ener} because the energy consumptions of their conditions are similar, but the moments of turning on the motor are different. It can be concluded from the trajectory that tacking with heading angle equal to $40^{\circ}$ costs the least energy in these trials, which saves 8.7-23.4\% energy compared to other conditions. The detailed data is presented in the Table \ref{energy}. The box plot in Fig. \ref{enpic}(a) analyzes the average and variance of the energy cost in each loop, and it also generates the same conclusion that $40^{\circ}$ is the optimal angle in terms of energy saving. \par

Fig. \ref{enpic}(b) shows the average energy consumption only for tacking. The results suggest that the smaller $\theta$ is, the less energy is needed to maneuver. Moreover, Fig. \ref{traj} shows that the blue, red and green trajectories contain 5 tackings while the cyan and pink trajectories contain only 4 tackings. It can be concluded that Hybrid Sailboat-II makes a longer tack when the heading angle is relatively small ($35^{\circ}$ and $40^{\circ}$). Based on our previous assumption, trajectories with $\theta$ equal to $35^{\circ}$ and $40^{\circ}$ would cost the least energy because their tacking distances are relatively long and they need less energy to maneuver. However, $\theta$ equal to $35^{\circ}$ takes 8.7\% more energy than $\theta$ equal to $40^{\circ}$. This is because $35^{\circ}$ is in the no-go zone of Hybrid Sailboat-II and the sailboat sometimes stops during sailing close-hauled. As a result, it takes longer time to finish the loop and the efficiency of sailing decreases, the total energy cost increases.\par

To sum up, the experiment shows the closer to the bound near no-go zone ($40^{\circ}$ in this experiment), the less energy the Hybrid Sailboat costs; on contrast, there is a trend for energy increasing when the heading angle increases ($45^{\circ}$, $50^{\circ}$  and $55^{\circ}$). Besides, if the heading angle is too small ($35^{\circ}$), which is in the no-go zone, the total cruising will cost even more energy. In this experiment, the best group saves 8.7 to 23.4 \% energy compared to other experiment groups and the improvement is quite considerable for ocean cruising. In addition, the lines shown in Fig. \ref{ener} is quite smooth thus their fitting curves can be used to predict the long-time energy cost at any time.

\begin{table}[thpb]
\caption{The Enengy Cost For Different Heading Angles}
\label{energy}
%\begin{center}
\centering
\begin{tabular}{|c|c|c|c|c|c|}
\hline
Heading angle&$35^{\circ}$&$40^{\circ}$&$45^{\circ}$&$50^{\circ}$&$55^{\circ}$\\
\hline
Total enenrgy cost(J) & 1275&1172&1344&1376&1446\\
\hline
Enenrgy cost(J) per loop &255&234&269&275&289\\
\hline
Tacking enenrgy cost(J) &11.70&12.02&12.34&14.87&16.66\\
\hline
\end{tabular}
%\end{center}
\end{table}

%-----------------------------------------------------------------------------------------
\section{Conclusions}

This paper addresses the problem of optimizing the energy consumption for autonomic cruising of Hybrid Sailboat-II. First, the mechatronic design of the Hybrid Sailboat is improved. Then the PID control for rudders is tested and the sail angle should be set as Fig. \ref{wforce} shows. In order to test the energy consumption of autonomic cruising, the trajectories are designed and real-time control strategy is provided. The experiment tests the relationship between energy consumption and heading angles $\theta$ during automatic cruising. The experiment results illustrate that $\theta$ do account for the energy consumption and the best $\theta$ can save up to 23.4\% energy than the worst $\theta$. The improvement is remarkable for automatic cruising, especially for long-time sailing. In addition, the fitting function of total energy consumption - time can be generated to predict future energy cost.\par
The relationship between the PWM values of the motor and turning efficiency has not been studied in this experiment, which will be our future topic. The best PWM value will be combined with the optimal heading angle $\theta$ to reduce more energy. Furthermore, the experiment is conducted on a calm pool with constant wind field, which is a pre-step to apply the Hybrid Sailboat to thr ocean cruising. Thus, increasing the robustness of current energy optimization method and applying it in the real environment are the further steps we are going to take.

%---------------------reference-------------------------------------------------------------

%-------------------------------------------------------------------------------------------
\end{document}